\documentclass[defaul,iicol]{sn-jnl}

\usepackage{graphicx}%
\usepackage{multirow}%
\usepackage{amsmath,amssymb,amsfonts}%
\usepackage{amsthm}%
\usepackage{mathrsfs}%
\usepackage[title]{appendix}%
\usepackage{xcolor}%
\usepackage{textcomp}%
\usepackage{booktabs}%
\usepackage{algorithm}%
\usepackage{algorithmicx}%
\usepackage{algpseudocode}%
\usepackage{listings}%
\usepackage{enumitem}
\usepackage{makecell}
\usepackage{multicol}
\usepackage{hyperref}

\theoremstyle{thmstyleone}%
\theoremstyle{thmstyletwo}%
\newtheorem{problem}{Problem}

\theoremstyle{thmstylethree}%
\allowdisplaybreaks
\DeclareMathOperator*{\argmax}{arg\,max}

\begin{document}


\title[ARMCHAIR: integrated inverse reinforcement learning and model predictive control for human-robot collaboration]{ARMCHAIR: integrated inverse reinforcement learning and model predictive control for human-robot collaboration}

\author*[1]{\fnm{Angelo} \sur{Caregnato-Neto}}\email{caregnato.neto@ieee.org}

\author[3]{\fnm{Luciano} \sur{Cavalcante Siebert}}\email{l.cavalcantesiebert@tudelft.nl }

\author[4]{\fnm{Arkady} \sur{Zgonnikov}}\email{a.zgonnikov@tudelft.nl }

\author[2]{\fnm{Marcos} \sur{R. O. A. Maximo}}\email{mmaximo@ita.br}

\author[1]{\fnm{Rubens} \sur{J. M. Afonso}}\email{rubensjm@ita.br}

\affil*[1]{\orgdiv{Electronic Engineering Division}, \orgname{Instituto Tecnol\'ogico de Aeron\'autica}, \orgaddress{\city{S\~ao Jos\'e dos Campos},  \state{S\~ao Paulo}, \country{Brazil}}}

\affil[2]{\orgdiv{Autonomous Computational Systems Lab (LAB-SCA), Computer Science Division}, \orgname{Instituto Tecnol\'ogico de Aeron\'autica}, \orgaddress{\city{S\~ao Jos\'e dos Campos},  \state{S\~ao Paulo}, \country{Brazil}}}

\affil[3]{\orgdiv{Intelligent Systems Department, Interactive Intelligence Group}, \orgname{Delft University of Technology}, \orgaddress{\city{Delft}, \country{The Netherlands}}}

\affil[4]{\orgdiv{Department of Cognitive Robotics}, \orgname{Delft University of Technology}, \orgaddress{\city{Delft}, \country{The Netherlands}}}
%
%
\abstract{
	One of the key issues in human-robot collaboration is the development of computational models that allow robots to predict and adapt to human behavior. 
	Much progress has been achieved in developing such models, as well as control techniques that address the autonomy problems of motion planning and decision-making in robotics. However, the integration of computational models of human behavior with such control techniques still poses a major challenge, resulting in a bottleneck for efficient collaborative human-robot teams.
	In this context, we present a novel architecture for human-robot collaboration: Adaptive Robot Motion for Collaboration with Humans using Adversarial Inverse Reinforcement learning (ARMCHAIR). Our solution leverages adversarial inverse reinforcement learning and model predictive control to compute optimal trajectories and decisions for a mobile multi-robot system that collaborates with a human in an exploration task. During the mission, ARMCHAIR operates without human intervention, autonomously identifying the necessity to support and acting accordingly. Our approach also explicitly addresses the network connectivity requirement of the human-robot team. Extensive simulation-based evaluations demonstrate that ARMCHAIR allows a group of robots to safely support a simulated human in an exploration scenario, preventing collisions and network disconnections, and improving the overall performance of the task.} 
\maketitle

\section{Introduction} \label{sec:intro}
%
A cyber-physical-human system (CPHS) is an emerging concept of control schemes that emphasizes the interaction between machines and humans \cite{roadmap2030}.
An exciting application of CPHS is the design of collaborative multi-robot systems (MRS) devised to support humans in a variety of tasks, from construction \cite{construction} to manufacturing \cite{manufacturing} and transport \cite{roadmap2030}. One of the core challenges in the conception of such MRS is the development of their motion planning and decision-making algorithms, which must enable the group of robots to harmoniously operate with humans towards an aligned objective \cite{survey_HRI_navigation}.

A particularly attractive technique used to this end is model predictive control (MPC), a receding horizon control scheme that has been extensively demonstrated for multi-robot real-time motion planning and decision-making \cite{AR2023,chanceConstraints,alonsoMora2021}. 
As a model-based approach, MPC typically requires accurate prediction models of the dynamic elements in the environment \cite{mpcbook}. This requirement is particularly relevant in the context of human-robot teaming, where appropriate group coordination depends on the capability of predicting and adapting to human decisions and motion in real-time \cite{LIU2023}.

The human behavior prediction models are not only a fundamental component of most predictive control schemes designed for human-robot collaboration but also outlined as a key element in the development of general CPHS \cite{roadmap2030}.
Traditionally, fields such as psychology rely on descriptive predictions of human behavior that are based on first principles \cite{roadmap2030,decisionFieldTheory2023,schumann2023using}. Alternatively, recent advances in machine learning techniques have enabled an increasing number of sophisticated data-driven models for human motion and decisions \cite{human_pred_model1,human_pred_model2}. As a quantitative methodology, machine learning offers the benefit of providing mathematical or algorithmic models that are naturally encoded into machines \cite{SurveyHumanModels2023}, facilitating their integration with powerful control techniques. As a result, machine-learned human behavior models and MPC have been successfully integrated to design efficient algorithms for CPHS \cite{alonsoMora2021,chanceConstraints}.

Another concrete motivation for this integration is the possibility of enforcing connectivity between the elements of the human-robot team. Often, proper operation of MRS requires the maintenance of network connectivity \cite{AR2023}.
This property holds particular importance when humans are integrated into the team since typical interactions, in the form of direct human commands, negotiations, or adaptations, also entail some form of communication. By  properly predicting human motion within the MPC framework through machine-learned behavior models, hard constraints on connectivity can be enforced, allowing the robots to receive commands, perform negotiations, or adapt to human actions by continually monitoring them. The maintenance of communication has also been demonstrated as an important factor in improving human-robot trust \cite{trustHRC2017}.
%
%
%


\subsection{Background}

Originally developed for process control in chemical plants \cite{cutler1980dynamic}, model predictive control (MPC) evolved over the last decades into a flexible tool that can be fine-tuned for a wide range of applications. In the field of robotics, it has been already demonstrated for motion planning of complex platforms such as autonomous cars \cite{chanceConstraints} and MRS \cite{Ferranti2023}. The inherent robustness associated with the closed-loop formulation alongside its capability to perform constrained optimization is ideal in addressing major issues in robot motion planning \cite{surveyMIP}, such as active collision avoidance in environments with moving obstacles \cite{alonsoMora2021}, safe operation in unknown environments \cite{raffo2019}, and coordinated rendezvous maneuvering  \cite{MRSRend}.
The use of MPC alongside mixed-integer programming (MIP) further enhances the technique, allowing the optimization of integer and continuous variables.  This extension allows MPC-MIP optimization models to encode a variety of additional requirements such as minimum-time maneuvering, task allocation, and connectivity maintenance \cite{variableHorizonMIP2002,Afonso2020,AR2023}.

To devise human prediction models, this work leverages inverse reinforcement learning (IRL) \cite{irlOriginal}, a machine learning method that has been extensively employed to model human behavior in the context of sequential decision-making, from aircraft pilots \cite{town2023pilot} to car drivers  \cite{IRL_driving}.
Unlike other popular machine learning methods used for this purpose, such as imitation learning \cite{survey_human_models}, IRL algorithms aim to recover a reward function given a dataset of demonstrations. 
%
This distinction offers substantial benefits since the ensuing reward functions can be used to infer human intentions and also provide enhanced flexibility, being used for optimizations considering scenarios that deviate from the ones present in the original data \cite{irlOriginal}. These traits have been leveraged for many applications, such as collaborative autonomous driving \cite{IRL_driving}, selecting trajectories for a manipulator robot that closely resemble observed human motion \cite{KoberIRL}, and inferring objectives of rogue drones \cite{IRL_infer_drone_obj}.

Adversarial Inverse Reinforcement Learning (AIRL) \cite{airl_original} is an alternate rendition of this method that leverages the principle of maximum-entropy IRL \cite{ziebart2008maximum} and adversarial deep networks. This approach has been shown to outperform established deep imitation learning algorithms in terms of generalization \cite{airl_original,sestini2021efficient}, demonstration-efficiency \cite{sestini2021efficient}, and precision \cite{maAIRL}. Similarly to traditional IRL, AIRL has found use in a multitude of applications, from the design of agents that can remain under human meaningful control \cite{Peschl_paper, cavalcante2023meaningful} to improving lane-changing \cite{AIRL_autonomousdriving} and lane-keeping maneuvers \cite{lee2022spatiotemporal} of autonomous vehicles. 

\subsection{Related work}
We start the presentation of the selected literature with works that employ MPC and machine learning to design algorithms for mobile human-robot teams.
In \cite{chanceConstraints}, an MPC-based formation controller was designed for human-leading truck platooning. A model of the driver's behavior was computed using IRL and then integrated into decentralized MPC controllers in each autonomous truck. The inherent uncertainty in the behavior of the drivers was handled using chance constraints.

The problem of coordinating a group of robots, humans, and moving obstacles without communication networks is addressed in \cite{alonsoMora2021}. A Recurrent Neural Network (RNN) is trained using a dataset of robot demonstrations to develop prediction models. The proposed MPC trajectory planners are able to steer independent robots in swap maneuvers without \textit{a priori} exchange of their intended paths.

A lane-changing problem in mixed-traffic scenarios is considered in \cite{mip_lanechanging}. The proposed approach relies on the recognition of six driver actions that are modeled using hidden Markov models. Probability models are then trained to predict the acceleration of the vehicle given the predicted human actions. An MPC-MIP formulation is employed to safely and efficiently steer the vehicle in the lane-changing maneuver. A strategy based on Interaction-Aware Model Predictive Path Integral (IA-MPPI) is proposed in \cite{jansma2023interactionaware} for trajectory planning of a team of Unmanned Surface Vessels (USV) in urban canals. The system is able to predict trajectories and goals using pedestrian models and social variational recurrent neural networks. 

In addition to MPC and machine learning methods, another popular methodology for the development of human-robot collaboration algorithms for mobile platforms are hierarchical architectures that leverage multiple distinct techniques that individually address the problems of decision-making, motion planning, and connectivity.
For example, \cite{hierarch1} proposes an architecture with three layers: a global planner leverages the fast marching square algorithm to provide proper trajectories to the goal, while a local planner addresses collision avoidance. A Social Force Model (SFM) with parameters estimated by a neural network is employed for human prediction. A Lloyd-based controller handles the multi-agent coordination problem and provides further safety guarantees.

A similar approach is proposed in \cite{kim2016socially}, where the task of path planning is divided into global and local planners, while collision avoidance is handled by a low-level controller. Human demonstrations are employed by an IRL algorithm to learn proper social navigation behaviors, which are then used to compute human-like trajectories for the robot.  A B-spline path generation algorithm is proposed in \cite{humanoid} to address the problem of humanoid robot navigation alongside humans. A combination of SFM and Bayesian human motion models is used to predict human trajectories. The integration between adaptive control strategies and social proximity potential fields has also been proposed as a solution for human-robot navigation under specific social conventions \cite{social_potential_fields}.

In \cite{semi_autonomous} a semi-autonomous approach is proposed where the human performs a search-and-rescue operation with the support of Unmanned Aerial Vehicles (UAVs). The tasks of the team are assigned by the operator while collision avoidance is handled by a trajectory planner based on replanning using linear spatial separations. Different levels of autonomy in human-robot teaming are studied in \cite{ship_docking} considering the task of robot-assisted docking of a ship conducted by a human.

The problem of human-robot collaborative search has been successfully demonstrated in \cite{IJSR_2023} through a series of thorough experiments. The proposal leverages the concept of Social Reward Sources (SRS) to model the environment and mission, while an architecture using Monte Carlo Tree Search (MCTS) and Rapidly-exploring Random Trees (RRT) is used to collaboratively navigate a group of robots. The interactions occur through messages with the humans transmitting their intended goals and positions using a smartphone.



\subsection{Contributions}

We propose Adaptive Robot Motion for Collaboration with Humans using Adversarial Inverse Reinforcement learning (ARMCHAIR), a novel architecture that leverages the integration between IRL and MPC-MIP for human-robot collaboration. Our work differentiates itself by:
\begin{enumerate}
	\item providing both movement coordination and collaboration in task allocation for an MRS that  supports a human in an exploration mission;
	\item not requiring human intervention, with the  robot team identifying the necessity of support and acting as needed;
	\item handling the joint decision-making and motion planning optimization problems in a unified MPC-MIP framework;
	\item actively enforcing network connectivity;
\end{enumerate}

Unlike most of the proposals found in the literature, ARMCHAIR addresses: a) the typical movement coordination problems associated with collision avoidance and connectivity maintenance that are required for proper coexistence and cooperation between humans and robots collaborating on the same task; and b) it provides harmonious task allocation with the machines updating their objectives to conform with human decisions, avoiding conflicts and effectively allowing the group to \textit{collaborate} towards an aligned global goal in spite of the uncertainty in the human's decisions. 

\begin{figure}
	\centering
	\includegraphics[width=0.45\textwidth]{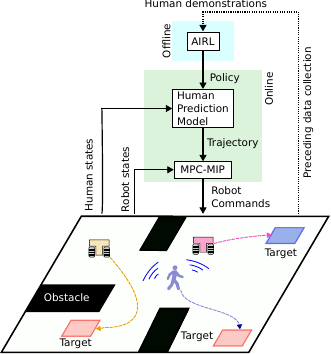}
	\caption{ARMCHAIR control architecture. The AIRL offline layer provides a human prediction model that is used by the MPC-MIP algorithm to compute proper trajectories and decisions for the MRS in a closed loop.}
	\label{fig:ctrl_arch}
\end{figure}

ARMCHAIR's architecture is presented in Figure \ref{fig:ctrl_arch}. The trajectories are computed with the proposed MPC-MIP scheme, which handles both the motion planning and task allocation issues in the same framework, being able to jointly optimize them to certified global optimality in finite-time \cite{richardsTutorial}. In addition, using the AIRL-based prediction model, ARMCHAIR is able to operate without human intervention by inferring their intentions and adapting to them. This is made possible, in part, by the active enforcement of network connectivity through the constraints in the MPC-MIP problem. The optimization is performed in a closed-loop fashion using the current feedback of the human and robot states, which are sensored and communicated to the planner through the network. The planner provides periodical updates on the trajectories and decisions of the MRS in a receding horizon fashion.

%



\subsection{Notation and  definitions}
The prediction of a variable ``$\circ$'' for the time step $k+\ell$ made at time step $k$ is written as $\hat{\circ}(\ell|k)$. Bold letters represent matrices and vectors, whereas scalars are written in plain text. The sets of all integers in the interval $[m,n]$ are written as $\mathcal{I}_m^n = \{m,m+1,\dots,n\}$. Sets are denoted by calligraphic letters as in $\mathcal{B}$. All polytopes mentioned in this work are convex. The Minkowski sum and Pontryagin difference operators are denoted by $\oplus$ and $\sim$, respectively \cite{Baotic2009}.


\section{Problem Description}\label{sec:II}

We consider a group of $n_a \in \mathbb{N}$ robots and one human in a task of visiting targets that represent regions of interest in a planar environment. This setup represents a general exploration mission, which is relevant in multiple real-world contexts such as search-and-rescue and surveillance tasks \cite{IJSR_2023}.
The discrete-time dynamics of the robots considering a sampling period of $T >0$ are described by the following equations
\begin{align}
	&\mathbf{x}_i(k+1) = \mathbf{A}_i\mathbf{x}_i(k) + \mathbf{B}_i\mathbf{u}_i(k) \label{eq:dyn}\\
	&\mathbf{y}_i(k) = \mathbf{Cx}_i(k) \label{eq:dyn_y},\ \forall i \in \mathcal{I}_1^{n_a}
\end{align}
where $\mathbf{x}_i \in \mathbb{R}^{n_x}$, $\mathbf{u}_i \in \mathbb{R}^{n_u}$, and $\mathbf{y}_i \in \mathbb{R}^2$ are the state, input, and position of robot $i$, respectively; $\mathbf{A}_i \in \mathbb{R}^{n_x \times n_x}$, $\mathbf{B}_i \in \mathbb{R}^{n_x \times n_u}$, and $\mathbf{C} \in \mathbb{R}^{2\times n_x}$ are the corresponding matrices of the state-space representation. States and inputs are bounded by the polytopic sets $\mathcal{X} \subset \mathbb{R}^{n_x}$ and $\mathcal{U} \subset \mathbb{R}^{n_u}$, respectively.

The group performs the task within the boundaries of a region represented by the polytope $\mathcal{A} \subset \mathbb{R}^2$ which contains $n_o \in \mathbb{N}$ polytopic obstacles $\mathcal{O}_g \subset \mathbb{R}^2$, $\forall g \in \mathcal{I}_1^{n_o}$. 
The space occupied by the body of the robots is defined as
\begin{align}
	\mathcal{R}_i(k) \triangleq \mathbf{y}_i(k) \oplus \mathcal{R}_i,\ \forall i \in \mathcal{I}_1^{n_a}
\end{align}
where $\mathcal{R}_i \subset \mathbb{R}^{2}$  is a polytope centered at the origin representing the shape of robot $i$. Similarly, the space occupied by the human is defined as
\begin{align}
	\mathcal{R}^h(k) \triangleq \mathbf{y}^h(k) \oplus \mathcal{R}^h
\end{align}
with $\mathbf{y}^h \in \mathbb{R}^2$ being the position of the human and $\mathcal{R}^h \subset \mathbb{R}^2$ representing the human body.
Define $\bar{\mathcal{O}} \triangleq \bigcup_{g=1}^{n_o} \mathcal{O}_g$. Then, the free operation region of the $i$-th robot at time step $k$ is written as, $\forall i \in \mathcal{I}_1^{n_a}$,
\begin{align}
	&\bar{\mathcal{A}}_i(k) = \left ( \mathcal{A} \setminus \left ( \bar{\mathcal{O}}\cup \mathcal{R}^h(k) \cup \bigcup_{\substack{j =1\\ j\neq i}}^{n_a}   \mathcal{R}_j(k)    \right) \right )  \sim \mathcal{R}_i
	\label{eq:freeSpace}
\end{align}

\begin{figure}
	\centering
	\includegraphics[width=0.42\textwidth]{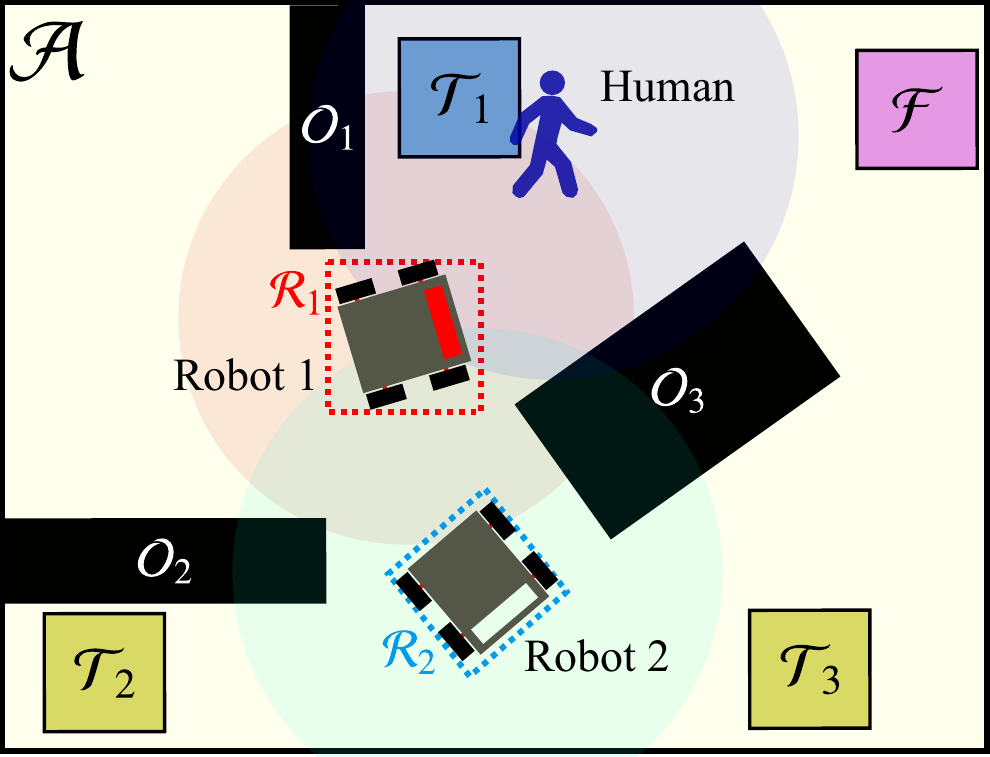}
	\caption{Illustration of environment with a target of type B in $\mathcal{T}_1$ and two targets of type A represented by the polytopes $\mathcal{T}_2$ and $\mathcal{T}_3$; three obstacles $\mathcal{O}_1$, $\mathcal{O}_2$, and $\mathcal{O}_3$; and a terminal region $\mathcal{F}$. The polytopes $\mathcal{R}_1$ and $\mathcal{R}_2$ are outer approximations of the robots' bodies. Connectivity regions are depicted as circles.}
	\label{fig:scenario}
\end{figure}

The targets in the environment are represented by the polytopes $\mathcal{T}_e \subset \mathbb{R}^2,\ \forall e \in \mathcal{I}_1^{n_t}$, where $n_t \in \mathbb{N}$ is the total number of targets. These regions are further divided into two types: A and B. From the perspective of the robots both types are equally valuable, whereas the human may be biased toward a particular type. In the design of the proposed algorithms, the preferences of the human are not explicitly known and can only be inferred from data. The number of targets of type A and B is denoted by $n_A \geq 0$ and $n_B \geq 0$, respectively, such that $n_A + n_B = n_t$.  The mission finishes when the human reaches a terminal region $\mathcal{F} \subset \mathbb{R}^2$. Figure \ref{fig:scenario} provides an example of the envisioned scenario.

In our setup, the group of robots must also maintain connectivity at each time step. We consider proximity-based links, where two agents are connected if they are close enough to each other, i.e., if their relative position is within a communication region represented by the polytope $\mathcal{C} \subset \mathbb{R}^2$. 
Under this condition for individual connections, the whole group can communicate if the underlying network, represented by an undirected graph, is connected \cite{Afonso2020}.

In order to cooperate with the human, the robots make use of a behavior model of that human. This behavior model is initially unknown, although the following assumptions are made: 
\begin{enumerate}[label=A{{\arabic*}})]
	\item a set of demonstrations $\mathcal{D}$ containing human trajectories in similar tasks is available;
	\item the human operates trying to maximize a compromise between visiting the maximum number of targets and finishing the mission as soon as possible. 
	\item the ensuing behavior is not necessarily optimal nor deterministic;
	\item the human moves with constant velocity.
\end{enumerate}

The set of human demonstrations assumed in A1 can be obtained by monitoring and collecting data directly during their activities. In this proof-of-concept work, we generated synthetic human behavior data rather than collecting it from real humans (see Appendix \ref{app:human_surrogate} for details).
 This approach allows for preliminary evaluation of algorithms for human-robot collaboration~\cite{Peschl_paper,uncertainty,sestini2021efficient,town2023pilot,jansma2023interactionaware} since it can generate human-like data in a scalable way, although it is limited in that it does not exactly represent actual human behavior. 
%
Assumptions A2 and A3 imply that the human is aware of their task and has some degree of competency in completing it. However, optimal performance is typically not achieved, as humans are susceptible to bounded rationality and uncertainty over objectives \cite{boundedRationality,biases,uncertainty}. We employed reinforcement learning to train a synthetic agent to emulate these characteristics (Appendix \ref{app:human_surrogate}). The resulting policy is optimized to maximize a compromise between collecting rewards (by visiting targets) and finishing the task quickly but is also stochastic and not necessarily globally optimal. Thus, both A2 and A3 are satisfied by the proposed synthetic agent.
For the sake of simplicity, Assumption A4 constrains the problem to humans moving with constant velocity. However, the methods herein proposed can be expanded to accommodate a varying velocity. 
In this context, the main problem addressed by ARMCHAIR is
\begin{problem}\label{prob:hrc}
	Design a motion planning and task allocation algorithm that allow the robots to collaborate with the human, complementing their efforts in the task when necessary without the necessity of human intervention.
\end{problem}


\section{Adaptive Robot Motion for Collaboration with Humans using Adversarial Inverse Reinforcement learning (ARMCHAIR)} \label{sec:III}
The ARMCHAIR scheme, presented in Figure \ref{fig:ctrl_arch}, addresses Problem \ref{prob:hrc} by leveraging AIRL to model human behavior through a dataset of their demonstrations while performing the mission in the environment.
The time-demanding learning process of AIRL is carried offline and only the resulting policy is used, alongside the feedback of current human and robot states, in the online closed-loop scheme. The MPC-MIP motion planning and decision-making algorithm performs an optimization considering the current state of the group and employs the policy computed by the AIRL method to predict future motion and decisions of the human given their current state. The inherent uncertainty in human actions is handled by a straightforward robustness approach using safety regions.

Each element of the ARMCHAIR's architecture is thoroughly discussed in the subsequent sections.

\subsection{Adversarial Inverse Reinforcement Learning}\label{sec:III-1}
Typically, MPC requires a model to predict the dynamic elements of the environment. Considering the problem description discussed in Section \ref{sec:II}, ARMCHAIR learns a prediction model for the human given only the set of demonstrations $\mathcal{D}$ under the aforementioned assumptions. 
To this end, we leverage AIRL \cite{airl_original} to recover a reward function so that ARMCHAIR can train a policy that imitates the behavior observed in the dataset. This policy will provide predictions of the human's high-level decisions, i.e., the targets that are likely to be visited, and the corresponding path through the environment. Henceforth, we followed the AIRL approaches from \cite{airl_original,Peschl_paper}.

In particular, we follow the scheme presented in \cite{Peschl_paper} and divide the environment presented in Section \ref{sec:II} into  $n_c \in \mathbb{N}$ cells of a grid. The ensuing scenario is represented by a Markov Decision Process (MDP) $\langle \mathcal{S},\mathcal{N},p,r,\gamma \rangle$ with 
\begin{align}
	\mathbf{s} = [\mathbf{F}_{\text{pos}},\mathbf{F}_\text{A},\mathbf{F}_\text{B},\mathbf{F}_{\text{obs}},\mathbf{F}_{\text{ter}}]
\end{align}
\begin{figure}[]
	\centering
	\includegraphics[width=0.3\textwidth]{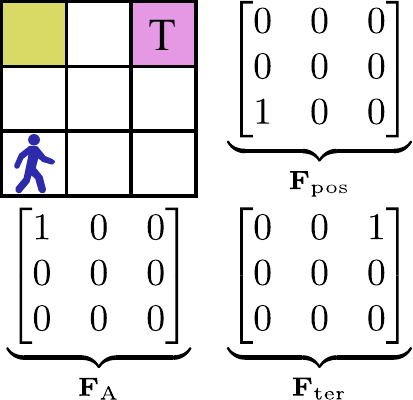}
	\caption{Human position, target of type A, and terminal region T in a grid encoded as the feature maps $\mathbf{F}_{\text{pos}}$, $\mathbf{F}_{\text{A}}$, $\mathbf{F}_{\text{ter}}$, respectively.}
	\label{fig:feature_maps}
\end{figure}
being a state belonging to the set of states $\mathcal{S}$ with the feature maps $\mathbf{F}_{\text{pos}},\mathbf{F}_\text{A},\mathbf{F}_\text{B},\mathbf{F}_{\text{obs}},\mathbf{F}_{\text{ter}}  \in \{0,1\}^{n_c \times n_c}$ representing the positions of the human, type A targets, type B targets, obstacles, and terminal region, respectively. Examples of feature maps are presented in Figure \ref{fig:feature_maps}. The set of $n_a \in \mathbb{N}$ actions is denoted as $\mathcal{N} = \{a_1,\dots,a_{n_a} \}$, whereas $p(\mathbf{s}' \vert \mathbf{s},a)$ is the transition distribution associated with the dynamics of the environment. 
The reward function and discount factor are denoted by $r: \mathcal{S} \times \mathcal{N} \rightarrow \mathbb{R}$ and $\gamma \in [0,1]$, respectively.

The two main components of the AIRL framework are the reward and policy networks represented by $f_\theta$ and $\pi_\phi$, which are parameterized by $\theta$ and $\phi$, respectively. The discriminator 
\begin{align}
	D_\theta(\mathbf{s},a,\mathbf{s}') = \dfrac{\exp(f_\theta(\mathbf{s},a,\mathbf{s}'))}{\exp(f_\theta(\mathbf{s},a,\mathbf{s}')) + \pi_\phi(a \vert \mathbf{s})}
\end{align}
previously proposed in \cite{airl_original} outputs the confidence that a particular state-action pair $(\mathbf{s},a)$ belongs to  the demonstration dataset rather than being generated from $\pi_\phi$. The discriminator is trained to improve this data classification using binary logistic regression, further refining the reward approximator $f_\theta$ at each iteration. The policy network is updated using PPO \cite{Peschl_paper}, to generate trajectories that better match the demonstrations, this way ``confusing" the classification process of the discriminator \cite{airl_original}. 
The policy recovered by AIRL is denoted by $\hat{\pi} : \mathcal{S} \rightarrow \Delta_{\mathcal{N}}$ and is a probability distribution over all actions conditioned by a state.

\subsection{Human Prediction Model}\label{subsec:human_pred_model}

Once $\hat{\pi}$ is properly learned, we can determine sequences of states representing the potential paths and decisions of the human given an initial state. However, due to the stochasticity of the policy, many distinct sequences can be computed. 
This uncertainty can be addressed by robust MPC planners with a multitude of techniques e.g., chance constraints \cite{chanceConstraints}, contingency \cite{contigencyMPC}, or constraint tightening. 
We opt for a straightforward approach where the predictions are the most likely sequence of states given by the policy, with the uncertainty in the human's actions being handled by the robust formulation presented in Section \ref{subsec:rob_form}. The sequence of states is computed through the propagation of the (\ref{eq:most_likely_act}) and (\ref{eq:most_likely_state}) until the terminal state is reached:

\begin{align}
	& \hat{a}(n) = \argmax_{a' \in \mathcal{N}} \hat{\pi}(a' \vert \hat{s}(n)) \label{eq:most_likely_act} \\   
	&\hat{\mathbf{s}}(n+1) = \argmax_{\mathbf{s}' \in \mathcal{S}} p(\mathbf{s}' \vert \hat{\mathbf{s}}(n),\hat{a}(n)) \label{eq:most_likely_state}
\end{align}
with $\hat{a}$ and $\hat{\mathbf{s}}$ representing action and state predictions, respectively. 

Algorithm \ref{algo: human_pred} summarizes the process of computing the human predictions. It starts by associating the first predicted state $\hat{\mathbf{s}}(0)$ with the measured state of the human at time step $k$, $\mathbf{s}^{\text{meas}}(k)$. The corresponding waypoint $\hat{\mathbf{w}}(n) \in \mathbb{R}^2 $, representing the predicted Cartesian position of the human, is then extracted from the feature map $\mathbf{F}_{\text{pos}}(n)$ using the function \textproc{getWaypoints}. 
Considering the current state, the procedures in lines 5 and 6 compute the most likely action and next state, respectively. This process is repeated in a loop until the predictions reach the terminal state.
Then, assuming that the human moves with a constant linear velocity $v > 0$, the predicted trajectories at time step $k$ are computed by the function \textproc{getTraj} yielding a sequence of predicted human positions, $\hat{\mathbf{y}}^h(\ell \vert k) \in \mathbb{R}^2,\ \forall \ell \in \mathcal{I}_0^{\bar{N}}$ with $\bar{N}$ being the maximum prediction horizon considered. 

Algorithm 1 runs periodically, employing updated measurements of the human's state, with a frequency dictated by the sampling period $T$.

\begin{algorithm}
	\caption{Human prediction model}
	\label{algo: human_pred}
	\begin{algorithmic}[1]
		\Require $\mathbf{s}^{\text{meas}}(k)$ \Comment{Human state at time step $k$}
		\Require $\hat{\pi}$  \Comment{Policy recovered with AIRL}
		\State $n \gets 0$
		\State $\hat{\mathbf{s}}(0) \gets \mathbf{s}^{\text{meas}}(k) $
		\While{ terminal region not reached } 
		\State $\hat{\mathbf{w}}(n) \gets$ \textproc{getWaypoints}($\hat{\mathbf{s}}(n)$)
		\State $\hat{a}(n) \gets \argmax_{a' \in \mathcal{N}} \hat{\pi}(a' \vert \hat{\mathbf{s}}(n))$
		\State $\hat{\mathbf{s}}(n+1) \gets \argmax_{\mathbf{s}' \in \mathcal{S}} p(\mathbf{s}' \vert \hat{\mathbf{s}}(n),\hat{a}(n))$
		\State $n \gets n + 1$
		\EndWhile
		\State $ \left \{\hat{\mathbf{y}}^h(\ell \vert k) \right \}_{\ell=0}^{\bar{N}} \gets$ \textproc{getTraj}$\left (\left \{\hat{\mathbf{w}}(j) \right \}_{j=0}^n \right)$
		\Ensure $\left \{\hat{\mathbf{y}}^h(\ell \vert k) \right \}_{\ell=0}^{\bar{N}} $
	\end{algorithmic}
\end{algorithm}

\subsection{MPC-MIP Formulation} 
The MPC-MIP formulation designed for the human-robot motion planning and decision-making problem is similar to \cite{AR2023}, with the distinction of accounting for an independent agent, i.e. the human, over which the group of robots has no command. Within the multi-agent team, the human is always assigned to the last index $n_a+1$, where $n_a$ is the number of robots, without loss of generality. Implementation details are available in the provided source code\footnote{\url{https://gitlab.com/caregnato_neto_open/ARMCHAIR}}.

Let the tuple $\boldsymbol{\lambda}= \left \langle \mathbf{x}_i,\mathbf{u}_i,b^{\text{hor}},b^{\text{tar}}_{i,e},b^{\text{con}}_{i,j} \right \rangle $ represent all variables to be optimized. Then, the human-robot teaming problem can be formalized as the following mixed-integer linear programming (MILP) model.
\begin{subequations}
	\allowdisplaybreaks
	\begin{flalign}
		&~\nonumber\\[10pt] 
		&\text{\textbf{Human-robot teaming optimization model}} \nonumber \\
		&\underset{\substack{\boldsymbol{\lambda}}}{\textrm{min}}   \sum_{k=0}^{\bar{N}} \left ( \sum_{i=1}^{n_a} \gamma_u\Vert \mathbf{u}_{i}(k) \Vert_1  - \gamma_t \sum_{e=1}^{n_t} \sum_{i=1}^{n_a+1} b^{\text{tar}}_{i,e}(k) \right)
		\label{eq:cost} \\
		&\textrm{s.t., } \nonumber\\
		& \textrm{\textbf{Human}} \nonumber \\
		& \forall \ell \in \mathcal{I}_0^{\bar{N}}, \nonumber \\
		& \mathbf{y}_{n_a+1}(\ell \vert k) = \hat{\mathbf{y}}^h(\ell \vert k) \label{const:human}\\
		& \textrm{\textbf{Robots}} \nonumber \\
		& \forall i \in \mathcal{I}_1^{n_a},\ \forall \ell \in \mathcal{I}_0^{\bar{N}} \nonumber\\
		%
		& \mathbf{x}_i(0 \vert k) = \mathbf{x}_i(k), \label{const:rec_hor}\\
		&\mathbf{x}_{i}(\ell+1 \vert k) = \mathbf{A}_i\mathbf{x}_i(\ell \vert k) + \mathbf{B}_i \mathbf{u}_{i}(\ell\vert k) \label{const:dynamics}\\
		& \mathbf{y}_i(\ell\vert k) = \mathbf{C}_i\mathbf{x}_i(\ell\vert k)\label{const:output} \\
		&\mathbf{x}_{i}(\ell+1\vert k) \in \mathcal{X}_i\label{const:state}\\
		&\mathbf{u}_{i}(\ell\vert k) \in \mathcal{U}_i \label{const:input}\\
		& \mathbf{y}_{i}(\ell+1\vert k) \in \bar{\mathcal{A}}_i(\ell+1\vert k) \label{const:freeSpace}\\
		& \textrm{\textbf{Collaboration}} \nonumber \\
		& \forall i \in \mathcal{I}_1^{n_a+1},\  \forall e \in \mathcal{I}_1^{n_t},\  i < j \leq n_a+1 \nonumber,\  \forall \ell \in \mathcal{I}_0^{\bar{N}} \label{const:tar_implication}\\
		& b^{\text{tar}}_{i,e}(\ell \vert k) \implies  \mathbf{y}_i(\ell \vert k ) \in \mathcal{T}_{e}\\
		& \sum_{k=0}^{\bar{N}} \sum_{h=1}^{n_a+1} b^{\text{tar}}_{i,e}(\ell \vert k) \leq 1 \label{const:max_reward} \\
		&  b^{\text{con}}_{i,j}(\ell \vert k)  \implies \mathbf{y}_i(\ell \vert k ) - \mathbf{y}_j(\ell \vert k )\in \mathcal{C}
		\label{const:comm_reg} \\
		& \text{deg}_i(\ell \vert k) = \sum_{\epsilon=1}^{i-1}b^{\text{con}}_{\epsilon,i}(\ell \vert k) + \sum_{\epsilon=i+1}^{n_a}b^{\text{con}}_{i,\epsilon}(\ell \vert k) \label{const:degree}\\
		&  \text{deg}_i(\ell \vert k) + \text{deg}_j(\ell \vert k) \geq  n_a,\ j\neq n_a+1
		\label{const:conn}\\
		&\sum_{i=1}^{n_a} b^{\text{con}}_{i,n_a+1}(\ell \vert k) \geq 1 \label{const:human_conn}\\
		& b^{\text{hor}}(\ell \vert k) \implies \mathbf{y}_{n_a+1}(\ell \vert k) \in \mathcal{F} \label{const:terminal}
	\end{flalign}
\end{subequations}

The cost (\ref{eq:cost}) is a compromise between the control effort of the robots (weighted by $\gamma_u >0$) and the collection of rewards through target visitation, with the rewards being represented by $\gamma_t >0$. Notice that the activation of the target binaries $b^{\text{tar}}_{i,e} \in \{0,1\}$, which implies that an agent visited a target due to constraint (\ref{const:tar_implication}), results in a discount that contributes to the minimization of (\ref{eq:cost}). Thus, the cost drives the group towards target visitation while also considering the total control effort (represented by the inputs $\mathbf{u}_i)$ of the maneuvers.

Constraint (\ref{const:human}) performs the assignment of the predictions coming from the human prediction model (Algorithm 1) to the appropriate variables in the optimization model, $\mathbf{y}_{n_a+1}$. 

Constraints (\ref{const:rec_hor}) - (\ref{const:input}) are related to the robots. Constraint (\ref{const:rec_hor}) establishes the receding horizon scheme of the MPC technique.  
The dynamics are encoded by (\ref{const:dynamics}) and (\ref{const:output}), whereas (\ref{const:state}) and (\ref{const:input}) enforce the robot's state and input bounds, respectively. 
Avoidance of collisions between elements of the group and with obstacles is enforced by (\ref{const:freeSpace}).

Finally, collaboration between the robots in the group is achieved through constraints (\ref{const:tar_implication}) - (\ref{const:terminal}). The relationship between target binaries, the positions of the agents, and the target polytopes is described by the implication in constraint (\ref{const:tar_implication}). Constraint (\ref{const:max_reward}) establishes that the reward associated with each target can be collected only once.
Constraint (\ref{const:comm_reg}) conditions the activation of the connectivity binary $b^{\text{con}}_{i,j} \in \{0,1 \}$ of agents $i$ and $j$ to their relative position being within the connectivity region polytope $\mathcal{C}$. The degree of the vertices that represent the agents in the communication network is encoded by (\ref{const:degree}). The condition for connectivity between robots is enforced by constraint (\ref{const:conn}), whereas (\ref{const:human_conn}) guarantees that there is always at least one robot connected to the human, forming a network with all elements of the team.

The mission is finished when the horizon binary $b^{\text{hor}} \in \{0,1 \}$ is active. Constraint (\ref{const:terminal}) conditions the activation of this binary to the human predicted position reaching the terminal region $\mathcal{F}$. 

\subsection{Robust formulation}\label{subsec:rob_form}

As discussed in \ref{subsec:human_pred_model}, the predictions employed by the motion planning algorithm take into account only the most likely actions of the human. We address the ensuing potential prediction errors by making the method more robust through an approach based on safety regions.

\begin{figure}[]
	\centering
	\includegraphics[width=0.2\textwidth]{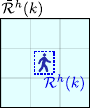}
	\caption{Safety region around the human to be avoided by the MRS.}
	\label{fig:safety_reg}
\end{figure}

To prevent potential collisions, we construct a safety region $\bar{\mathcal{R}}^h \subset \mathbb{R}^2$, that must be avoided by the robots, by enlarging the polytope representing the body of the human to account for any possible direction of movement at the next step (Figure \ref{fig:safety_reg}). 
Define, $\forall i \in \mathcal{I}_1^{n_a}$, $\forall \ell \in \mathcal{I}_0^{\bar{N}}$,
\begin{align} 
	& \mathcal{P}_i(k+1) \triangleq \bar{\mathcal{O}}\cup \bar{\mathcal{R}}^{h}(k) \cup \bigcup_{\substack{j =1\\ j\neq i}}^{n_a}   \mathcal{R}_j(k+1)   
	\label{eq:not_allowed}
\end{align}
as the union of the sets representing the obstacles, the body of the human, and the bodies of robots for $i\neq j$, i.e., the regions that robot $i$ is not allowed to occupy over time. Then, we adapt the free operation region (\ref{eq:freeSpace}) used in constraint (\ref{const:freeSpace}) as, $\forall i \in \mathcal{I}_1^{n_a}$, $\forall k \in \mathcal{I}_0^{\bar{N}}$,
\begin{align}
	&\bar{\mathcal{A}}_i(\ell+1\vert k) = 
	( \mathcal{A} \setminus \mathcal{P}_i(\ell+1 \vert k )) \sim \mathcal{R}_i
	\label{eq:freeSpace_exp}
\end{align}

A similar approach is employed to prevent network disconnections. A shrunken connectivity region is computed considering the safety region
\begin{align}
	\bar{\mathcal{C}} = \mathcal{C} \sim \bar{\mathcal{R}}^h
\end{align}
and constraint (\ref{const:comm_reg}) is rewritten to employ the shrunk region when considering connectivity with the human, $\forall i \in \mathcal{I}_1^{n_a}$, $\forall \ell \in \mathcal{I}_0^{\bar{N}-1}$, $j = n_a+1$,
\begin{align}
	b^{\text{con}}_{i,j}(\ell+1 \vert k)  \implies \mathbf{y}_i(\ell \vert k ) - \mathbf{y}_j(\ell \vert k )\in \bar{\mathcal{C}}
\end{align}


\section{Results} \label{sec:IV}
We evaluated the proposed scheme considering two environment configurations with four obstacles, one terminal region, and a maximum of four targets. Environment 1 was designed to investigate the functionality of ARMCHAIR in a situation where the targets are sparsely distributed, whereas environment 2 contains closely grouped targets that represent a more challenging problem in terms of human behavior prediction and appropriate robot response.

In both scenarios, the simulated human is supported by a group of $n_a=2$ robots with discrete-time double integrator dynamics described by (\ref{eq:dyn}) and (\ref{eq:dyn_y}) considering a sampling period of $T=1$ s. The states are $\mathbf{x}_i = [r_{x,i},v_{x,i},r_{y,i},v_{y,i}]^\top,\ \forall i \in \mathcal{I}_1^{n_a}$, where $r_{x,i}$ and $r_{y,i}$ are the positions of robot $i$ in the $x$ and $y$ axes, respectively, and $v_{x,i}$ and $v_{y,i}$ are the corresponding velocities; the input $\mathbf{u}_i = [a_{x,i}, a_{y,i}]^\top,\ \forall i \in \mathcal{I}_1^{n_a}$, is comprised of the corresponding accelerations. Although simplified, this choice of dynamics model for motion planning has been successfully demonstrated in experiments with real robots using MPC-MIP setups \cite{AR2023}. The velocity and acceleration of the robots are bounded to the intervals $[0,2.5]$ m/s and $[-5,5]$ m/s$^2$, respectively. A circular connectivity region with radius $4$ m is approximated by a regular octagonal polytope inscribed in the circle $\mathcal{C} \subset \mathbb{R}^2$, following \cite{AR2023}. The width and length of the robots, human, and operation region are $0.5 \times 0.5$ m, $0.41 \times 0.29$ m, and $7 \times 7$ m, respectively. The weights of (\ref{eq:cost}) were selected as $\gamma_u = 0.01$ and $\gamma_t = 5,\ \forall t \in \mathcal{I}_1^4$, i.e., all targets are equally valuable for the robot team.

As stated in Section \ref{sec:II}, we employed a synthetic agent to represent the human, which operates independently from the robots. Its use allows for a deeper evaluation of the proposed scheme as data can be generated to study the method in a multitude of circumstances. The actions of the synthetic agent are determined by a policy $\pi^*: \mathcal{S} \rightarrow \Delta_{\mathcal{N}}$. The dataset $\mathcal{D}$ used in the AIRL training process discussed in Section \ref{sec:III-1} is generated using $\pi^*$. 

The effectiveness of ARMCHAIR was evaluated by comparing it to two baselines: 
\begin{enumerate}[label=\alph*)]
	%
	\item open-loop MIP: An open-loop version of our algorithm, where the optimization is solved once at the start of the mission considering the most likely trajectory of the human given by $\hat{\pi}$;
	\item No robot support: The human performs the task without robot support.
\end{enumerate}

A comparison to the baseline without any robot support allowed us to evaluate the influence of proper robot support on the task. Furthermore, by comparing ARMCHAIR with an open-loop MIP approach, we aimed to investigate the impact of the closed-loop strategy employed by the former method, where trajectories and decisions are periodically updated, as opposed to the latter, where a single optimization is performed at the start of the mission.

Since the actions of the simulated human are stochastic, we performed a Monte Carlo evaluation considering 1000 simulations in each environment. To illustrate the performance of ARMCHAIR compared to the baseline methods in each environment, we first highlight several representative simulations in detail (recordings of all 1000 simulations are available in the supplementary material\footnote{\url{https://gitlab.com/caregnato_neto_open/ARMCHAIR}}). We then present statistical analysis across all simulations considering the following metrics: average number of collisions, network disconnections, targets visited, and redundant target visitations, where a target is visited twice in the same simulation due to a conflict in target allocation. 

\subsection{Environment 1: Sparse target distribution}

\begin{figure}[h!]
	\centering
	\includegraphics[width=0.45\textwidth]{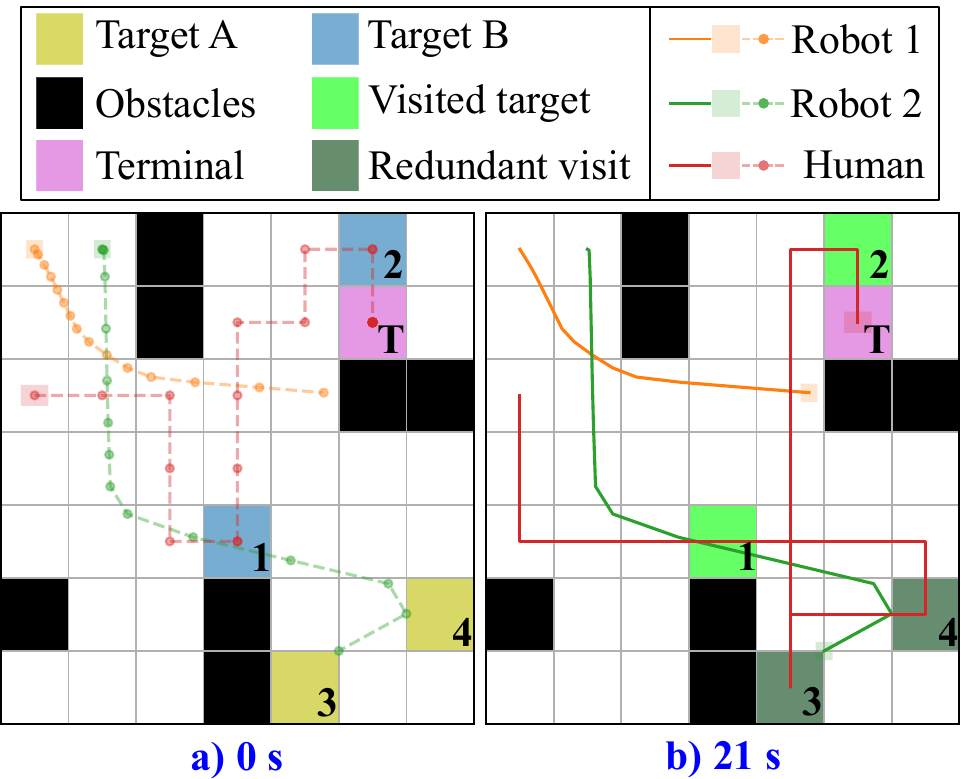}
	\caption{Example simulation of the open-loop MIP baseline in Environment 1 (Sparse target distribution): initial a) and final b) time steps for simulation 2 out of 1000. Solid and dashed lines represent the actual motion and predictions of each agent, respectively. The human prediction is computed using only the initial conditions (open-loop) and $\hat{\pi}$. The prediction suggests that only targets of type B will be visited by the human and the robots are dispatched to the remaining ones. In b) we observe redundant visits to targets 3 and 4 (Type A) since the human deviates from the prediction. }
	\label{fig:result_env1_mip}
\end{figure}



\begin{figure*}[]
	\includegraphics[width=1\textwidth]{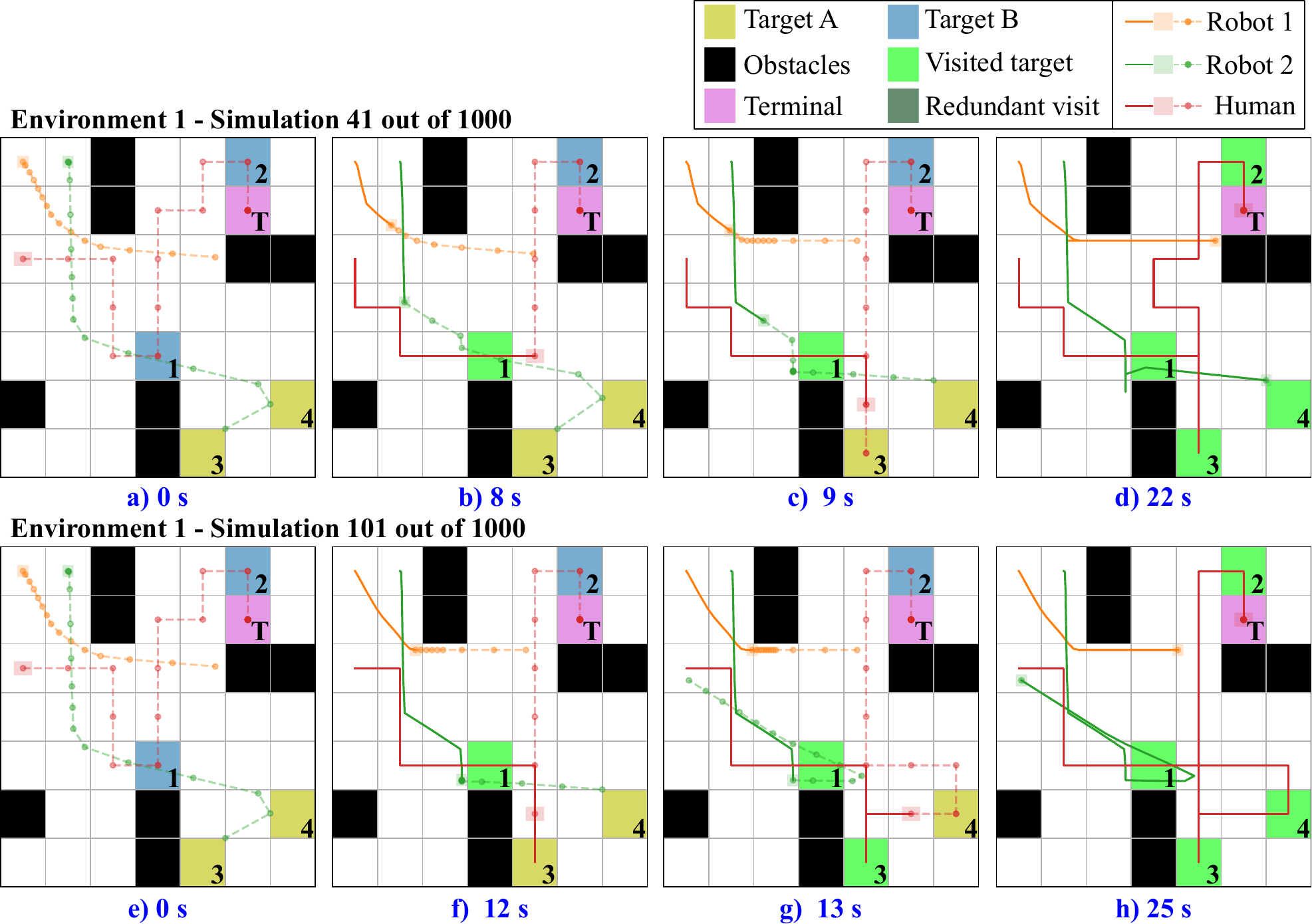}
	\caption{Example simulation of ARMCHAIR in Environment 1 (Sparse target distribution): Four time steps of simulations 41 and 111 out of 1000. Solid and dashed lines represent the motion and predictions of each agent, respectively.  The ARMCHAIR algorithm allows the replanning of the robot's trajectories and target assignments, preventing any redundant visits when the human deviates from the initial prediction.}
	\label{fig:result_env1_mpc}
\end{figure*}

\begin{table*}[h!]
	\caption{Performance of ARMCHAIR compared to the baseline methods in Environment 1 (Sparse target distribution). Each metric is represented by mean and 95\% bootstrap confidence intervals (CI) for 1000 simulations.}
	\centering
	\begin{tabular}{lccccc}
		\toprule
		& & Collisions & Disconnections & Targets & Redundant visits     \\
		\midrule
		ARMCHAIR   & \makecell{Mean\\CI} & \makecell{$\mathbf{0.001}$\\ $(0.0,0.003)$} & \makecell{$\mathbf{0.0}$\\ $(0.0,0.0)$} & \makecell{$\mathbf{4.0}$\\$(4.0,4.0)$} & \makecell{$\mathbf{0.0}$\\ $(0.0,0.0)$}  \\ 
		Open-loop MIP & \makecell{Mean\\CI} & \makecell{$0.58$\\$(0.53,0.63)$} & \makecell{$0.02$\\$(0.01,0.03)$} & \makecell{$4.0$\\$(4.0,4.0)$} & \makecell{$0.43$\\$(0.40,0.47)$} \\ 
		No robot support& \makecell{Mean\\CI}& N$\backslash$A & N$\backslash$A & \makecell{$2.7$\\$(2.6,2.7)$} & N$\backslash$A  \\
		\bottomrule
	\end{tabular}\label{tab:results_env1}
\end{table*}

We found that in the sparse environment, open-loop MIP produces frequent redundant visits to targets in cases when the human deviated from the initially most likely course of action. For instance, in simulation 2 out of 1000 at the initial time step (Figure~\ref{fig:result_env1_mip}a) the learned policy $\hat{\pi}$ predicted a trajectory where the human visits only the preferred targets 1 and 2, ignoring the remaining ones. In response, the open-loop MIP dispatched robot 2 to visit targets 3 and 4, while robot 1  maintained the connectivity between the human and robot 2. However, after 21 s (Figure \ref{fig:result_env1_mip}b) the human drifted from the initial prediction by visiting all targets, resulting in redundant visits of targets 3 and 4 which were reached both by the human and robot 2.

Conversely,  we observed that the redundant visits were completely prevented when the agents operated using the ARMCHAIR algorithm, as illustrated by simulations 41 and 111 out of 1000 in Figures \ref{fig:result_env1_mpc}a to \ref{fig:result_env1_mpc}h. 
Simulation 41,  shows a significant detour (Figure \ref{fig:result_env1_mpc}b and \ref{fig:result_env1_mpc}c) where the human demonstrates its intention to visit target 3 by moving downwards instead of upwards. ARMCHAIR promptly updates the trajectories of robot 1, stopping it for a few time steps allowing the human to move away from the region around target 4, and then dispatching robot 2 to visit it. Simulation 101 depicts a similar situation, where the human visits target 3 but also decides to capture target 4 (Figures \ref{fig:result_env1_mpc}f and \ref{fig:result_env1_mpc}g). As a result, ARMCHAIR updates the trajectory of robot 2, steering it away from the bottom right-hand side region, clearing the path for the human to finish the mission while robot 1 preserves the connectivity of the group.

Confirming the insights based on representative simulations, ARMCHAIR demonstrated superior performance compared to both baselines across all 1000 simulations (Table \ref{tab:results_env1}). In particular, ARMCHAIR provided proper responses to the uncertainty in human behavior, as no network disconnections or redundant visits occurred, whereas such issues were present in the open-loop MIP case. The average number of redundant visits was particularly high (0.43) for open-loop MIP, due to the incapability of the open-loop planner to update trajectories and allocation of tasks in real-time. ARMCHAIR also produced few collisions (on average 0.001 collisions per simulation) compared to the open-loop MIP (0.58 collisions per simulation) which was unable to adapt to the human deviating from the initially estimated trajectory. In terms of efficiency, all targets were visited when the open-loop and ARMCHAIR planners were used, as opposed to the case where the human operated alone (2.7). However, only ARMCHAIR prevented redundant visits.

\subsection{Environment 2: Grouped target distribution}

\begin{figure}[]
	\centering
	\includegraphics[width=0.45\textwidth]{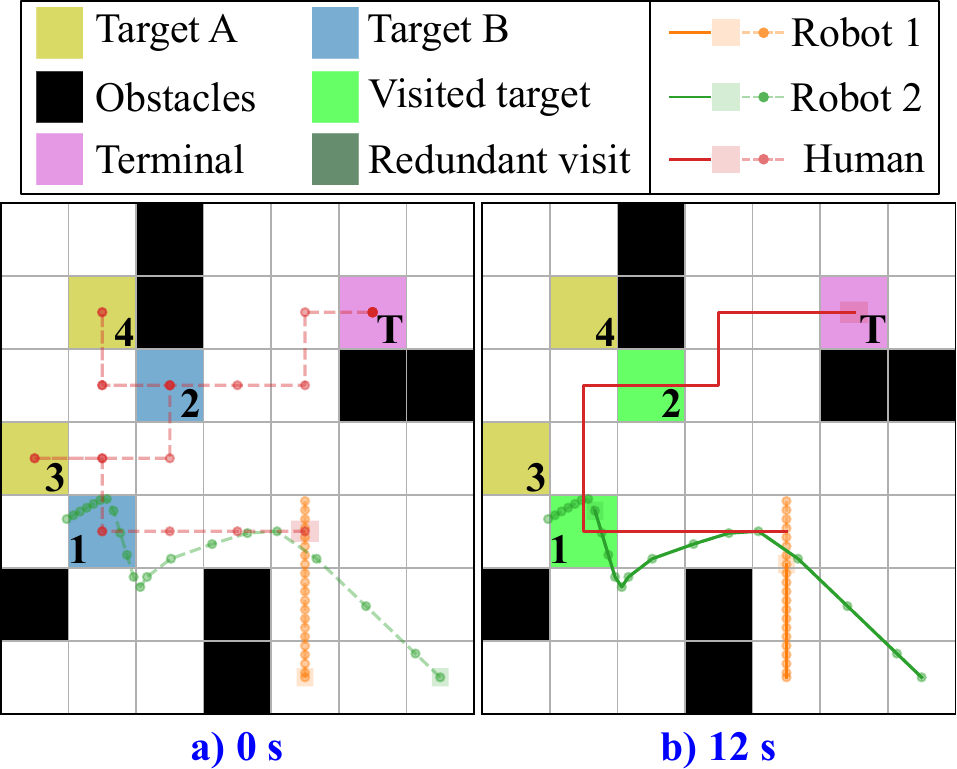}
	\caption{Example simulation of the open-loop MIP baseline in Environment 2 (Grouped target distribution): initial a) and final b) time steps for simulation 1 out of 1000. Solid and dashed lines represent the actual motion and predictions of each agent, respectively. The human prediction is computed using only the initial conditions (open-loop) and $\hat{\pi}$. The prediction suggests that all targets will be visited by the human and the robots are dispatched only to maintain connectivity. In b) we observe that the human decides to ignore the targets of type A deviating from the initial prediction. As a result, the human-robot team provides inferior performance in the mission.}
	\label{fig:result_env2_mip}
\end{figure}

In the environment with grouped targets, the initial prediction from the learned policy (Figure \ref{fig:result_env2_mip}a) indicates that the human is likely to visit all targets in the environment since they are close to each other. Consequently, the robots are dispatched only to maintain connectivity. After 12 seconds (Figure \ref{fig:result_env2_mip}b) we observed that the human deviates from the predictions, choosing to visit only targets 1 and 2  and finish the mission by reaching the terminal region. This result illustrates a distinct issue with the open-loop MIP approach: if in the initial prediction all targets are assigned to be likely visited by the human and a deviation occurs, the robot team is unable to update its decisions and motion to compensate for unexpected human behavior, yielding worse performance in the mission.

%

%

\begin{figure*}[h]
	\includegraphics[width=1\textwidth]{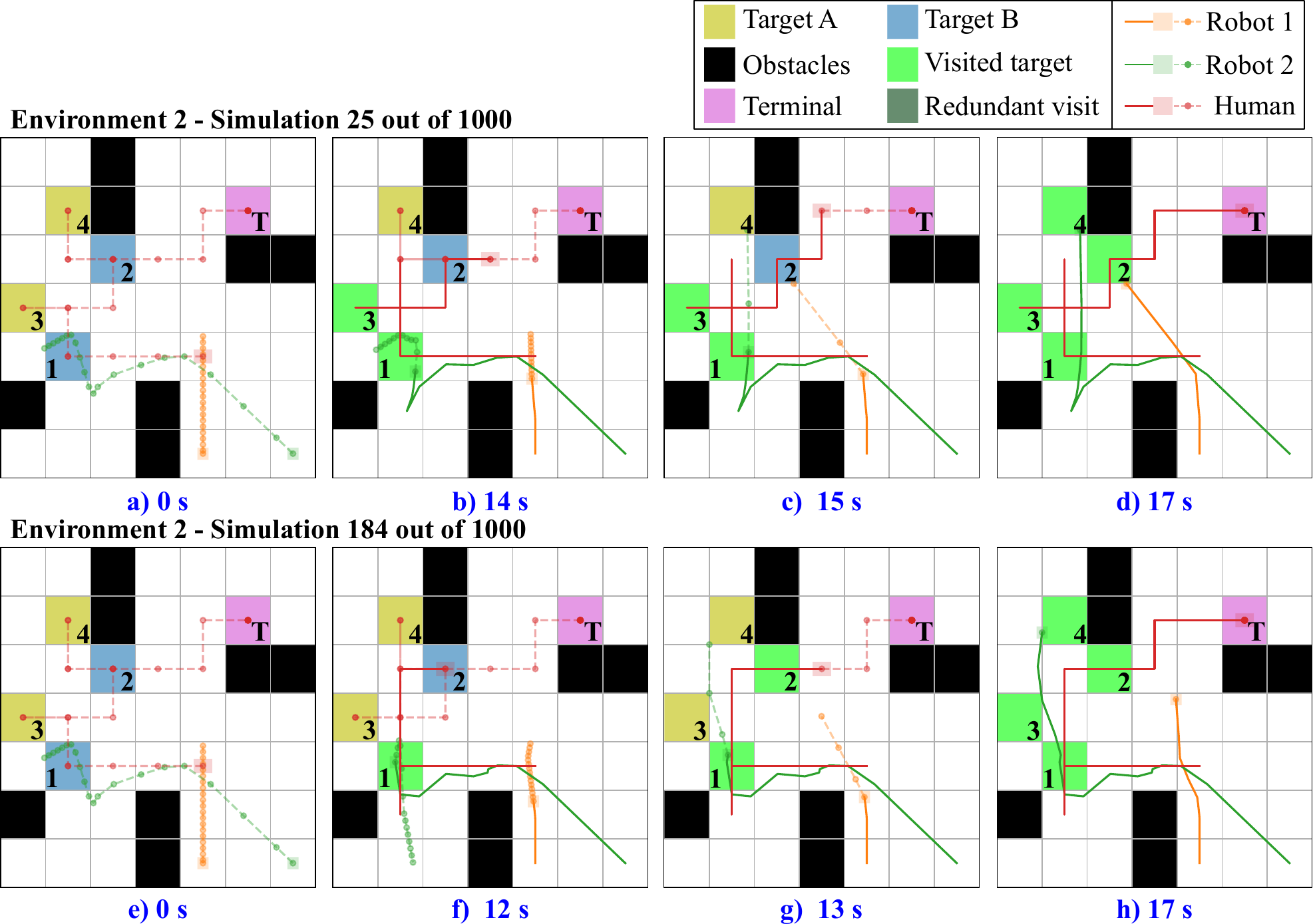}
	\caption{Example simulation of ARMCHAIR in Environment 2 (Grouped target distribution): Four time steps of simulations 25 and 184 out of 1000. Solid and dashed lines represent the motion and predictions of each agent, respectively. In this case, the replanning capabilities provided by the ARMCHAIR algorithm allow the robots to quickly adapt to compensate for an unexpected change in the behavior of the human, who decides to ignore targets that are easily accessible. As a result, all targets are visited at the end of the mission.}
	\label{fig:result_env2}
\end{figure*}

\begin{table*}[]
	\caption{Performance of ARMCHAIR compared to the baseline methods in Environment 2 (Grouped target distribution). Each metric is represented by mean and 95\% bootstrap confidence intervals (CI) for 1000 simulations. }
	\centering
	\begin{tabular}{lccccc}
		\toprule
		& & Collisions & Disconnections & Targets & Redundant visits     \\
		\midrule
		ARMCHAIR  & \makecell{Mean\\CI} & \makecell{$0.013$\\ $(0.007,0.02)$} & \makecell{$0.0$\\ $(0.0,0.0)$} & \makecell{$3.444$\\ $(3.398,3.489)$} & \makecell{$0.0$\\ $(0.0,0.0)$}  \\ 
		Open-loop MIP & \makecell{Mean\\CI} & \makecell{$0.004$\\$(0.001,0.008)$} & \makecell{$0.181$\\$(0.157,0.205)$} & \makecell{$2.384$\\$(2.354,2.415)$} & \makecell{$0.0$\\$(0.0,0.0)$} \\ 
		No robot support  & \makecell{Mean\\CI}& N$\backslash$A & N$\backslash$A & \makecell{$2.393$\\$(2.362,2.424)$} & N$\backslash$A  \\
		\bottomrule
	\end{tabular}\label{tab:results_env2}
\end{table*}

The simulation results in Figure \ref{fig:result_env2} show how ARMCHAIR is able to enhance the performance in the mission even when the human behavior predictions are imperfect. In simulation 25 out of 1000, the prediction that the human will visit all targets is maintained until time step 14 (Figure \ref{fig:result_env2}b), and the robots operate very similarly to the open-loop MIP case. Unexpectedly, the human decides not to capture target 2 even though it was properly positioned to do so. By moving upwards at time step 15 (Figure \ref{fig:result_env2}c) the human indirectly declares its intention to finish the mission without visiting the remaining targets, and the predictions are readily updated with the human going directly towards the terminal cell. Even with this unexpected behavior of the human and the small window of time remaining in the mission, ARMCHAIR was able to update trajectories for robots 1 and 2 such that targets 4 and 2 were visited (Figure \ref{fig:result_env2}d).

In simulation 184 out of 1000, the human presented a more conventional behavior, visiting the preferred targets of type B (Figure \ref{fig:result_env2}f) but again deviating from the initial prediction. Figure \ref{fig:result_env2}g shows that after visiting target 2, the human moves towards the terminal cell and the predictions are updated accordingly. ARMCHAIR  reacts by planning a trajectory for robot 2 in which it visits the remaining targets 3 and 4, while robot 1 maintains connectivity.

These observations are supported by the statistical results presented in Table \ref{tab:results_env2}. As in the first environment, ARMCHAIR was able to maintain the network connected in all simulations, whereas an average of 0.181 disconnections occurred when the open-loop MIP scheme was used.
Although no redundant visits were registered with either approach, ARMCHAIR provides a higher average number of target reward collections (3.444) than the open-loop algorithm (2.384). Better performance in terms of average reward collection is achieved with the robot team is used as opposed to the case without support, for which the average was (2.393). 

We also observed that ARMCHAIR performed worse in terms of collision avoidance in environment 2. This result stems from the increased complexity of the motion planning problem in this scenario due to the tight placement of targets, which requires the group to operate in close proximity.
Furthermore, the results indicate that the open-loop MIP scheme outperformed ARMCHAIR in terms of collision avoidance in this particular environment. This occurs because the robot team is not dispatched to visit any targets when the open-loop approach is employed. As a result, the robots  typically remain far away from the human and collisions are less likely. Conversely, ARMCHAIR consistently updates the trajectories and decisions of the robots to improve performance. Consequently, they are occasionally steered towards targets closer to the human, increasing the likelihood of collisions.


\section{Discussion and Conclusion} \label{sec:V}

This paper presented ARMCHAIR, a novel architecture for human-robot collaboration that leverages the integration of AIRL and MPC. Through extensive Monte-Carlo simulation trials, we demonstrated that the scheme provides proper adaptive trajectories and decision-making, in the form of target allocations, for a group of robots supporting a human in an exploration mission. 

ARMCHAIR was successful in both: a) coordinating the motion of the robot team, preventing most collisions and network disconnections; and b) autonomously identifying when to provide support based on the observed human behavior and current mission circumstances. This was made possible by the adaptive decision-making provided by the MPC-MIP scheme, which improved performance by preventing overlaps in the allocation of targets to robots and the human.

The achieved results illustrate that ARMCHAIR provides distinct capabilities compared to most of the literature on human-robot collaboration with mobile robots. Excellent solutions for social navigation (coordination) between robots and humans are found in \cite{chanceConstraints,alonsoMora2021,mip_lanechanging,social_potential_fields}. However, the scenarios considered in this paper require the extra capability of harmonious collaboration towards the same global objective. The ensuing joint problem of human-robot trajectory planning and decision-making has been successfully addressed in \cite{semi_autonomous} and \cite{IJSR_2023}, with the latter work thoroughly demonstrating its effectiveness through experiments with humans. However, both of these solutions are still dependent on frequent human inputs. In \cite{IJSR_2023}, for example, the necessity of the human operator to divide their attention between sending messages to the robot team and working on the mission was reported as an open issue. ARMCHAIR addresses this gap by actively predicting and adapting to the operator based solely on the observed human behavior. Thus, our approach has an extra degree of autonomy that allows the operator to concentrate on performing the mission, without the necessity of explicitly directing the actions of the robot team. 

Due to the novelty of the proposed integration, ARMCHAIR is still bound by a few limitations. For example, its receding horizon formulation lacks a formal demonstration of recursive feasibility and has no active method to handle unfeasibility, simply commanding the robots to stop moving until the problem becomes feasible again. In addition, as a centralized scheme, our approach may have scalability issues when larger teams are considered. The simulation-based evaluations presented illustrate that the integration of MPC-MIP and AIRL for human-robot teaming algorithms is promising. However, further investigation with thorough experimental demonstrations involving real humans, as performed in \cite{IJSR_2023} considering similar exploration missions, is still required.

In future works, the investigation of conditions for formal demonstrations of theoretical properties, in particular recursive feasibility, should be addressed. Reactive strategies, such as inner potential field control laws, could be devised to handle the occasional collisions that occur due to infeasible optimization problems.
The human prediction model can be refined to address more complex concepts, such as intentions and visual perceptions; the modeling of the human motion could also be improved, allowing for smoother movement predictions that would reduce the conservatism in the MRS trajectories.
Alternative methodologies for robustness could also be explored, reducing conservatism by tolerating risk. Distributed MIP schemes could be investigated to address the scalability issues.
The use of demonstrations with direct interaction between robots and humans can provide enhanced capabilities to the human prediction model. Experimental trials using real humans and robots can further validate the effectiveness of the proposal.

Overall, this work demonstrates that the integration of AIRL and MPC-MIP in ARMCHAIR yields a platform that already provides notable capabilities for human-robot collaboration and can be further refined to become an enabling technology for safe, harmonious, and efficient human-robot teaming in the future.

\bmhead{Supplementary information} 

All the algorithms and resulting data related to this paper are available at the repository\footnote{\url{https://gitlab.com/caregnato_neto_open/ARMCHAIR}}.

\bmhead{Author Contributions}

ACN devised the research question, performed the literature review, prepared all figures and tables, wrote the new code required, devised the simulation environments, evaluated results, and wrote all sections. LCS, AZ, MROAM, and RJMA provided support in determining the research question and performing the literature review, suggested the addition of figures, helped in the determination of simulation environments and evaluation methodology, and contributed to the writing and proofreading of all sections.

\bmhead{Funding}

This study was financed in part by the Coordena\c c\~ao de Aperfei\c coamento de Pessoal de N\' ivel Superior - Brasil (CAPES) - Finance Code 001. Marcos Maximo is partially funded by CNPq -- National Research Council of Brazil through the grant 307525/2022-8. The participation of Luciano Cavalcante Siebert and Arkady Zgonnokov in this research was partially supported by TAILOR, a project funded by EU Horizon 2020 research and innovation programme under GA No 952215.

\section*{Declarations}

\bmhead{Conflict of interest}

The authors declare that they have no conflict of interest or competing financial interests or personal relationships that could have appeared to influence the work reported in this paper.

\begin{appendices}
	
	\section{Synthetic human agent}
	\label{app:human_surrogate}
	As a surrogate for the human, we employ a reinforcement learning agent trained using Proximal Policy Approximation (PPO) on the ground truth rewards detailed in Table \ref{tab:rewards}. Its actions represent movement in the \textit{north}, \textit{south}, \textit{west}, and \textit{east} directions, as well as a \textit{collection} that allows the agent to capture target rewards. The relative value of the targets to the human is represented by the rewards $R_A > 0$ and $R_B > 0$, with $R_B > R_A$. Thus, it has a preference for the type B. The rewards can only be collected at most once. There are three types of penalization: collisions with walls and obstacles, movement, and failure in reaching the terminal region in the required number of steps, they are represented by the quantities $R_C < 0$, $R_M < 0$, and $R_F <0$, respectively.
	
	\begin{table}
		\centering
		\caption{Ground-truth rewards and penalizations for the synthetic human agent.}
		\begin{tabular}{ccccccc}\toprule
			& $R_A$ & $R_B$ & $R_C$ & $R_M$ & $R_F$ & \\
			Value & 0.5    & 1.0    & -1.0  & -0.1 & -20 & \\
			\bottomrule
			\label{tab:rewards}
		\end{tabular}
		
	\end{table}
	
	We employ the same PPO setup as in \cite{Peschl_paper} with $\gamma = 0.999$.
	The training employs new randomly generated environments similar to the one depicted in Figure \ref{fig:result_env1_mpc}. For each environment, the maximum number of targets is $n_t \sim \mathcal{U}(0,4)$. The number of target of type $B$ and $A$ is computed as $n_{B} \sim \mathcal{U(}0,n_t)$ and $n_{A} = n_t - n_{B}$, respectively. The initial position is also randomized for each scenario, being uniformly distributed over all cells that are not targets or obstacles. The position of the obstacles and terminal region remains fixed. The human does not consider the robot's support explicitly but rather reacts indirectly, through their impact on the environment, e.g., capture of targets.

	After running the algorithm for $16\times10^5$ steps, the synthetic agent's policy $\pi^*: \mathcal{S} \rightarrow \Delta_\mathcal{N}$ is computed, where $\Delta_\mathcal{A}$ is a probability distribution over all actions. Finally, $\pi^*$ is used to generate a set of $8000$ trajectories, $\mathcal{D}$, representing the human navigation in different environments. 
	
	\section{AIRL training}
	\label{app:AIRL}
	
	Table \ref{tab:airl_pars} presents the hyperparameters used in the AIRL training. We evaluate the performance of the recovered policy by comparing the trajectories generated by $\hat{\pi}$ and $\pi^*$ in 1000 new environments built randomly using the same procedure discussed in Appendix A. Table \ref{tab:airl_results} summarizes the results in terms of average length of the trajectory, number of targets of type A and B visited, collisions, and return over the ground truth rewards considering the trajectories achieved using the synthetic human agent and the recovered policies.
	\begin{table}
		\centering
		\caption{AIRL hyperparameters.}
		\begin{tabular}{ccc}\toprule
			& Hyperparameter & Value \\
			\midrule
			& Learn. rate discr. & $4e{-4}$ \\
			& Learn. rate gen. & $4e{-4}$ \\
			& Batch size discr. & 256 \\
			& Batch size gen. & 8 \\
			& Environment steps & $3.5e6$ \\
			& $\epsilon$-clip & 0.1 \\
			& $\gamma$ & 0.999 \\
			& PPO epochs & 5 \\
			\bottomrule
			\label{tab:airl_pars}
		\end{tabular}
	\end{table}
	
	The results show that the policy recovered by the AIRL algorithm generates useful predictions of the human agent. The resulting average trajectory lengths are very similar, suggesting that the prediction model accurately captures the ``urgency" of the human in finishing the mission. The obstacle avoidance capability is also appropriately emulated with no collisions occurring in both cases. However, there are mismatches in the target collection averages, in particular targets of type B. The recovered model shows a less greedy behavior, with smaller target collection averages. 
	
	\begin{table}
		\centering
		\caption{Comparison between recovered and surrogate human policies.}
		\begin{tabular}{lcc}\toprule
			\textbf{Synthetic human agent} & &  \\
			\midrule 
			& Average & Confidence Interval \\
			Trajectory length  & 12.7    & $(12.3,13.0)$ \\
			Tar. A collection & 0.68    & $(0.63,0.73)$ \\
			Tar. B collection & 1.09    & $(1.02, 1.16)$ \\
			Collisions & 0.0    & $(0.00,0.00)$ \\
			Return & 0.05    & $(0.01, 0.09)$ \\
			\midrule
			\textbf{AIRL recovered} &   & \\
			\midrule 
			Trajectory length  & 12.3    & $(11.9,12.7)$ \\
			Tar. A collection & 0.61    & $(0.56,0.66)$ \\
			Tar. B collection & 0.81    & $(0.75,0.87)$ \\
			Collisions & 0.0    & $(0.00,0.00)$ \\
			Return & -0.85    & $-(1.01,0.70)$ \\
			\bottomrule
			\label{tab:airl_results}
		\end{tabular}
	\end{table}

	
	
	
	
	
	
\end{appendices}

\bibliographystyle{sn-nature} 
\bibliography{sn-bibliography}

\end{document}